\documentclass[10pt, a4paper]{article}
\usepackage{lrec2022} % this is the new LREC2022 Style
\usepackage{multibib}
\newcites{languageresource}{Language Resources}
\usepackage{graphicx}
\usepackage{tabularx}
\usepackage{soul}
\usepackage{titlesec}
\titleformat{\section}{\normalfont\large\bfseries\center}{\thesection.}{1em}{}
\titleformat{\subsection}{\normalfont\SmallTitleFont\bfseries\raggedright}{\thesubsection.}{1em}{}
\titleformat{\subsubsection}{\normalfont\normalsize\bfseries\raggedright}{\thesubsubsection.}{1em}{}
\renewcommand\thesection{\arabic{section}}
\renewcommand\thesubsection{\thesection.\arabic{subsection}}
\renewcommand\thesubsubsection{\thesubsection.\arabic{subsubsection}}
\usepackage{epstopdf}
\usepackage[utf8]{inputenc}

\usepackage{hyperref}
\usepackage{xstring}

\usepackage{color}

\title{Life is not Always Depressing: Exploring the Happy Moments of People Diagnosed with Depression}

\name{Ana-Maria Bucur$^{1,2}$, Adrian Cosma$^{3}$, Liviu P. Dinu$^{1}$} 

\address{$^1$ University of Bucharest, Romania $^2$ Universitat Politècnica de València, Spain \\
$^3$ University Politehnica of Bucharest, Romania\\
         ana-maria.bucur@drd.unibuc.ro, cosma.i.adrian@gmail.com, ldinu@fmi.unibuc.ro}

\abstract{
In this work, we explore the relationship between depression and manifestations of happiness in social media. While the majority of works surrounding depression focus on symptoms, psychological research shows that there is a strong link between seeking happiness and being diagnosed with depression. We make use of Positive-Unlabeled learning paradigm to automatically extract happy moments from social media posts of both controls and users diagnosed with depression, and qualitatively analyze them with linguistic tools such as LIWC and keyness information. We show that the life of depressed individuals is not always bleak, with positive events related to friends and family being more noteworthy to their lives compared to the more mundane happy events reported by control users.
\\ \newline }

\begin{document}

\maketitleabstract

\section{Introduction}
\textit{"I have the choice of being constantly active and happy or introspectively passive and sad. Or I can go mad by ricocheting in between"}, wrote the famous poet Sylvia Plath in her personal diaries \cite{plath_hughes_2000}. It is now known that Sylvia Plath was clinically depressed most of her life, before tragically committing suicide. Depression is the modern ailment of our times. It is reported that more than 264 million people of all ages suffer from this mental illness \footnote{\url{https://www.who.int/news-room/fact-sheets/detail/depression}}. More than 700 thousand people take their own life every year, and even more people attempt suicide \footnote{\url{https://www.who.int/news-room/fact-sheets/detail/suicide}}. Thankfully, modern psychology has a fair understanding of depression, its onset, its symptoms and potential treatments, more recently aided by computational methods \cite{orabi2018deep,burdisso2019text,uban2020deep,inkpen2021uottawa,bucur2020detecting,DBLP:journals/corr/abs-2106-16175}. However, research in automatic understanding of manifestations of moments of happiness is lacking. How much do we understand about the happy moments of depressed individuals? In this preliminary work, we aim to bridge this gap and develop a computational method for extracting and analyzing happy moments from a large corpus of social media text.

Indeed, promoting happiness is an important aspect of our lives, and is one of the main ways in which we can prevent disease and live a longer, healthier life. While our happiness is determined largely by genetic factors \cite{diener1999subjective}, it can be influenced in a significant manner through our behavior: long-term relationships with friends and family, meaningful work and self-care. Paradoxically, \newcite{Ford2014} showed that individuals in cultures that highly value happiness are less likely to attain it. Moreover, the authors show that highly valuing happiness is a key indicator and a potential risk factor of depression. In the same vein, \newcite{ALARCON2013821} found that happiness is positively related to optimism and hope and that these traits are negatively associated with depression and anxiety. Recent computational tools \cite{DBLP:conf/emnlp/CarageaDD18,iordache} currently allow for accurate modelling of the optimism-pessimism continuum from text, but very little research is performed on the specific instances of happy moments, and on what aspect makes an individual happy. We posit that while both depressed and control subjects experience happy moments, the reasons for their happiness differ. 

In this preliminary work, we explore through linguistic tools the qualitative difference of happy moments between individuals diagnosed with depression and controls in social media text. While the majority of work in the area of exploring manifestations of depression in text focuses on specific symptoms \cite{inkpen2021uottawa,bucur2020detecting,DBLP:journals/corr/abs-2106-16175}, we focus our attention on exploring happiness manifestations in both depressed individuals and control. For this goal, we make use of HappyDB \cite{asai-etal-2018-happydb}, a database of 100.000 expressions of happy moments, and conduct our analysis on eRisk \cite{losada2016test}, a popular dataset of social media posts from Reddit for early detection of mental disorders. Different from sentiment polarity, which determines the sentimental aspect of an opinion, happiness is a broader concept that incorporates well-being and psychological health, referring to a positive affective state. Since there is no available dataset for modeling happy moments in depressed individuals and control, we use the Positive-Unlabeled learning paradigm to automatically extract happy moments from users' posts, by combining HappyDB with eRisk to obtain a Positive-Unlabeled dataset. We train an SVM in this paradigm \cite{elkan2008learning} on top of BERT \cite{devlin-etal-2019-bert} embeddings to extract happy moments, and use LIWC \cite{pennebaker2001linguistic} and keyness statistics to make a qualitative exploration of the happy moments. This work paves the way to a more in depth understanding of depression through computational methods. In this way, automatic tools can be developed to aid in early diagnosis, therapeutic chatbots, and can provide more insights for developing suitable positive psychological interventions (PPIs) \cite{parks2016positive} (e.g., feeling and expressing gratitude, pursuing meaning, building hope) for people diagnosed with depression. PPIs have been shown to improve physical and mental health in patients with symptoms of depression \cite{d2015happiness}.

Through our analysis, we show that happy moments differ in people diagnosed with depression compared to control subjects, users from the depression group are more likely to consider happy moments those events related to family and friends. Our analysis shows that, for users from the control group, the main reasons for happiness are related to entertainment situations and financial plans.

This paper has the following contributions:

\begin{enumerate}
    \item We propose a method for automatically extracting happy moments from social media text using the Positive-Unlabeled learning paradigm.
    \item We make a qualitative exploration of happy moments present in the Reddit posts of both depressed individuals and controls and found noteworthy differences between the two groups, which are confirmed by psychological studies.
\end{enumerate}

\section{Related Work}
In psychology, there are different scales for measuring the happiness of an individual, such as The Oxford Happiness Questionnaire \cite{hills2002oxford}, The Satisfaction with Life Scale \cite{diener1985satisfaction}, The PNAS Scales \cite{watson1988development} and others. The PNAS Scales measures different feelings or emotions (e.g., enthusiastic, alert, nervous, excited) in specific periods of time (e.g., in this moment, today, past few days), while The Oxford Happiness Questionnaire and  The Satisfaction with Life Scale are comprised of different statements on happiness, unhappiness and life satisfaction.

In the field of computational linguistics, \newcite{mihalcea2006corpus} proposed a corpus-based approach to finding happiness. The paper analyzes a corpus of blog posts from the LiveJournal community annotated with sad and happy emotions. The authors' findings reveal some patterns of happiness and sadness: 3 pm and 9-10 pm are the happiest hours, and Wednesday is the saddest day of the week. After analyzing the words from different semantic classes (e.g., actions, things, socialness), the authors conclude the work with a recipe of happiness with the following ingredients: "something new", "lots of food that you enjoy", "your favorite drink" and "an interesting social place".

In the realm of understanding manifestations of happiness, most notably, the Workshop on Affective Content Analysis proposed to tackle the problem of modeling happy moments in the CL-Aff Shared Task 2019 – in Pursuit of Happiness \cite{jaidka2019cl}. It had two tasks: a) predicting the Agency (if the author is in control, if the happy moment is influenced entirely by their actions) and Sociality (if other people besides the author are involved in the happy moment) of happy moments and b) suggesting insights  to characterize the happy moments automatically. 

Even if many works exploring the social media discourse of people diagnosed with depression \cite{orabi2018deep,burdisso2019text,uban2020deep,bucur-etal-2021-exploratory} are paying attention to the emotions expressed in their social media discourse \cite{aragon2021detecting,lara2021deep,howes2014linguistic}, to the best of our knowledge, works focusing on happiness felt by individuals diagnosed with depression are missing. Therefore, we aim to fill this gap and explore the happy moments from the online discourse of users with depression in comparison with control users.

\section{Data}
\begin{figure*}[hbt!]
    \centering
    \includegraphics[width=0.90\textwidth]{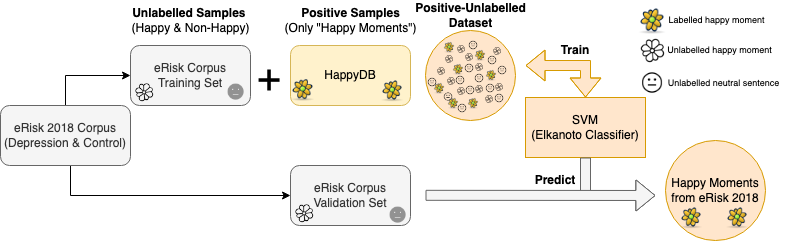}
    \caption{Positive-Unlabeled training paradigm, applied to extracting happy moments from eRisk depression dataset. We randomly sample 50,000 sentences from eRisk (unlabeled positive and negative classes) and 50,000 sentences from HappyDB (labeled positive class) to construct a Positive-Unlabeled dataset. After encoding the sentences with BERT, we trained a linear SVM in the PU-learning paradigm and use it to extract happy moments from eRisk validation set.}
    \label{fig:method}
\end{figure*}

In our experiments, we use two datasets comprised of English texts: HappyDB and the eRisk 2018 depression dataset.

\textbf{HappyDB} \cite{asai-etal-2018-happydb} is a corpus of 100,000 happy moments, comprised of sentences with crowdworkers' responses to the questions \textit{"what made you happy in the past 24 hours?"} or \textit{"what made you happy in the past 3 months?"}. It is a diverse corpus, containing happy moments from a variety of topics such as family (e.g., \textit{"My daughter waved at me and said 'mama' for the first time when I came home this morning."}), food (e.g., \textit{"I ate a juicy piece of pizza."}), work (e.g., \textit{"It made me happy to get a promotion at work."}), entertainment (e.g., \textit{"I went to a concert with friends."}) and others.

In this paper, we use the cleaned version of the HappyDB, in which spelling errors were corrected and empty or single-word sentences were removed. An important aspect of HappyDB is that it contains only positive examples (i.e., statements of happiness) and does not contain any negative examples, which makes it difficult to use in standard classification settings.

\textbf{The eRisk 2018 depression dataset} \cite{losada2016test} contains posts from Reddit and it was used for the early risk detection of depression task from the Early risk prediction on the Internet (eRisk) Workshop\footnote{\url{https://early.irlab.org/}}. Users were annotated as having depression by their mention of diagnosis (e.g., \textit{"I was diagnosed with depression"}) in their posts. Expression such as \textit{"I have depression"} or \textit{"I am depressed"} were not taken into account in annotating the users, only users with explicit mentions of depression diagnosis were labeled as having depression. Users from the control group are random users who do not have any mention of diagnosis in their posts, some of them may be active in the depression subreddit\footnote{\url{https://www.reddit.com/r/depression/}}. The posts disclosing the depression diagnosis were removed by the authors from the dataset. The datasets contains 214 users diagnosed with depression and 1493 control users with approximately 90,000 submissions from the depression group and over 985,000 submissions from the control group.

\section{Methods}
To conduct our analysis, we first trained an SVM model on top of BERT sentence embeddings that is able to extract happy moments from social media text. We then analyzed the happy moments using LIWC \cite{pennebaker2001linguistic} and keyness information.

\subsection{Extracting Happy Moments}
In order to extract relevant happy moments from the eRisk dataset, we train a classifier according to the Positive-Unlabeled Learning framework \cite{elkan2008learning}. In this setting, the goal is to train a binary classifier to distinguish between positive classes (i.e., the happy moments) and negative classes (i.e., any other type of text). However, during classifier training, only some of the positive examples are labeled, and none of the negative examples are labeled. Formally, we have a dataset of triples $(x, y, s)$, where $x$ is an example, $y$ is the class label, $y \in \{0,1\}$, and $s$ is a variable representing whether the pair $(x, y)$ is labeled or not. As such, if the example is labeled, it belongs to the positive class: $P(y=1|s=1) = 1$. When variable $s=0$, the pair can belong to either class, 0 or 1 in our case. We chose to adopt the PU-learning methodology, as opposed to a strictly unsupervised approach, to include additional information in the computation from the eRisk dataset, which contains Reddit posts qualitatively different from crowd-sourced texts from HappyDB. 

In our case, we split the posts from the eRisk dataset into sentences, and construct a training dataset comprised of approximately 50,000 sentences from HappyDB \cite{asai-etal-2018-happydb} and 50,000 sentences from the posts of the first 10 subjects from eRisk (from both depression and control classes). This dataset is a Positive-Unlabeled dataset (PU dataset), since the sentences from HappyDB are always the positive class, but the sentences from eRisk can contain both happy moments and neutral (non-happy) text. This methodology is presented in Figure \ref{fig:method}. Due to limited computational resources, we choose to use only a subset of the datasets for training our model. We use the rest of the subjects as a validation set to conduct our analysis. 

We train a linear SVM \cite{cortes1995support} with high regularization using the Elkanoto \newcite{elkan2008learning} training methodology on the PU dataset. We extract semantic features using a pretrained SentenceBERT \cite{devlin-etal-2019-bert} model and train the SVM on the sentence representations. We used the \textit{pulearn}\footnote{\url{https://github.com/pulearn/pulearn}} python library to train our classifier. To conduct our analysis, we used the previously trained SVM model on our eRisk validation set. Examples of extracted happy moments from eRisk validation set are presented in Table \ref{tab:erisk-happy}.

{\def\arraystretch{1.5}
\begin{table}[hbt!]
    \centering
    \resizebox{0.5\textwidth}{!}{
    \begin{tabular}{p{0.3\textwidth} | p{0.3\textwidth}}
        \textbf{Depression} & \textbf{Control} \\
        \hline
         Just found my first Rush dream setlist from long ago. & Yesterday morning, we woke up to a remarkably pleasant surprise. \\
         I just ordered two pairs of these...they look super comfy! & I visited Japan recently.\\
         I lost 120lbs through diet and exercise. & This new vegan supermarket just opened near me!\\
         Getting under 200 is the best feeling! & I'm on vacation and there's this amazing yarn shop nearby!\\
         Did get a novella published with a small press, working on bigger stuff. & I recently quit smoking weed. \\
         I decided that I had to leave the job and make my new job the task of trying to improve my situation. & About a month ago, my dad finally went into alcohol rehab.\\
         I recently adopted a cat and he's settled into his new home. & We went to some amazing gigs and hung out.\\
         My sister bought me The Last of Us Remastered and I've just been grinding through it. & I get all excited when I receive an actual letter from someone I personally know.\\
    \end{tabular}
    }
    \caption{Selected happy moments automatically extracted from our validation set of eRisk dataset using Positive-Unlabeled learning. Different from sentiment polarity, happy moments express significant moments of positive affective states.}
\label{tab:erisk-happy}
\end{table}
}

\subsection{Exploring Happiness}

For exploring the happiness moments expressed in the eRisk 2018 dataset, we use the Linguistic inquiry and word count (LIWC) lexicon \cite{pennebaker2001linguistic} due to its psychologically meaningful categories. The LIWC 2001 lexicon contains words from several categories such as linguistic dimensions (e.g., pronouns, articles, negations), psychological processes (e.g., affective processes, cognitive processes), and personal concerns (e.g. occupation, leisure). In our analysis, we aim to find the origin of people's happy moments, thus we analyze only the suitable categories from LIWC; we are not interested in the linguistics dimensions. As such, we chose the categories of sensory and perceptual processes (e.g. HEARING, FEELING), social processes (e.g., FAMILY, FRIENDS), and personal concerns (e.g., SCHOOL, SPORTS, RELIGION).

For comparing the happy moments from the two corpora (depression and control), we compute the dominance scores \cite{mihalcea2009linguistic} for the LIWC features analyzed, thus comparing the relative difference in the use of words from different categories between users diagnosed with depression and control users. The coverage is computed by comparing the words from the foreground corpus (F) with the words from the background corpus (B).
The coverage of a specific class (C) in a corpus (X) is computed as follows:

\begin{equation}
    Coverage_{X}(C) = \frac{\sum_{w_{i}\epsilon C} Frequency_{X}(W_{i})}{Size_{X}}
\end{equation}

where $C = W_{1}, W_{2}, ..., W_{N}$ represents a class of words.

The dominance score of the foreground corpus with respect to the background corpus is computed as a ratio between the coverages of the two corpora:

\begin{equation}
    Dominance_{F}(C) = \frac{Coverage_{F}(C)}{Coverage_{B}(C)}
\end{equation}

A dominance score equal or close to 1 for a given class indicates that the foreground and background corpora contain a similar distribution of words from the class. A score greater than 1 indicates that the foreground corpus contains more words from a given class than the background corpus. In contrast, a value lower than 1 suggests that words from a class are underrepresented in the foreground corpus instead of the background corpus.

\begin{table*}[hbt!]
    \centering
    % \resizebox{\textwidth}{!}{
    \begin{tabular}{ccp{4cm}|ccp{4cm}}
        \multicolumn{3}{c}\textbf{Depression} & \multicolumn{3}{c}\textbf{Control} \\
        \hline
        Category & Score & Example words & Category & Score & Example words \\
        \hline
        FRIEND S & 1.81 & friend, love, boyfriend, girlfriend, mate, bud, neighbor & TV & 1.74 & ad, show, movie, tv, actor, video\\
        MUSIC & 1.48 & song, sing, band, rap, listen, music & BODY & 1.21 & eye, heal, breath, pregnant, cheek, skin\\
        SLEEP & 1.47 & night, bed, sleep, dream, woke, nap & MONEY & 1.41 & fee, owe, rent, bought, pay, invest\\
        FAMILY & 1.45 & dad, kin, ex, son, great, family, mom & DEATH & 1.02 & die, dead, decay, dying, ashes, terminate\\
        SEXUAL & 1.38 & sex, bi, love, stud, fuck, hug & RELIGION & 0.99 & sin, christ, hell, god, angel, bless\\
        SELF & 1.36 & I, me, my, we, us, our, mine & SPORTS & 0.97 & play, game, ski, team, running, sport, exercise\\
    \end{tabular}
    % }
    \caption{Top classes and dominance scores for depression and control. Example words for each category are the ones found in our two corpora. }
    \label{tab:dominance}
\end{table*}

\section{Results and Discussion}
We computed dominance scores with each class in the foreground to reveal the dominant LIWC categories in the two corpora (depression and control). In Table \ref{tab:dominance} we have on the left side the analysis with the happy moments from the depression class as the foreground corpus and the happy moments of control subjects as the background corpus. On the right side of Table \ref{tab:dominance}, the moments from the control users are the foreground corpus. From these results, it is apparent that depressed individuals more frequently manifest happy moments relating to family and friends (dominance scores $>$ 1.4): \textit{"I've been training regularly with my entire family for 3 months now and I love every bit of it."}, \textit{"I'm with a woman I love and I have really supportive friends and family."}. Such events are more likely to be considered of greater value when compared to the happy moments of the control group. The control group, however, more frequently exhibit happy moments in everyday situations, in contexts related to entertainment, sports and financial plans: \textit{"Knowing I make more money than you does make me happy."}, \textit{"Eating pop corn and seeing movie."}, \textit{"I bought a new controller and worked perfect."}. In line with psychology research, \newcite{10.1371/journal.pone.0153715} found that the support of family and friends is negatively correlated with depression symptoms, in a study conducted on adolescents. Events related to positive experiences with loved ones are more noteworthy in a depressed subject's life, more so than in the case of control subjects.
\begin{figure}[hbt!]
    \centering
    \includegraphics[width=0.72\linewidth]{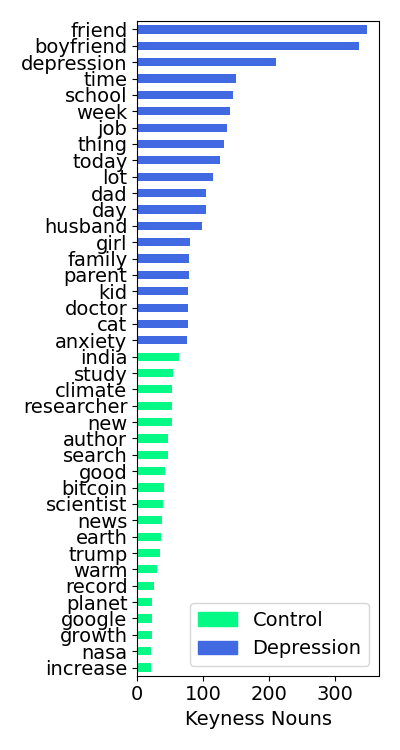}
    \caption{Keyness scores for nouns in the extracted happy moments from eRisk. Individuals with depression are more likely to discuss subjects related to family and friends than control users.}
    \label{fig:keyness-nouns}
\end{figure}

\begin{figure}[hbt!]
    \centering
    \includegraphics[width=0.72\linewidth]{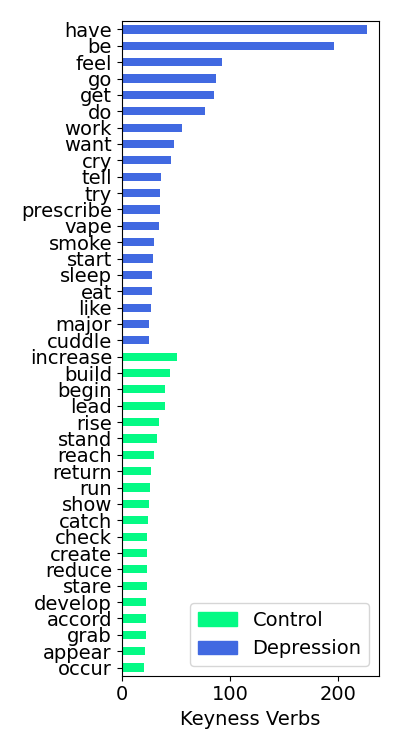}
    \caption{Keyness scores for verbs in the extracted happy moments from eRisk. Individuals with depression are more likely to use emotion-related verbs, while users from the control group use more action verbs.}
    \label{fig:keyness-verbs}
\end{figure}

Moreover, introspective discourse is more prevalent in depressed individuals (i.e., SELF category), which is in accordance with other works \cite{DBLP:journals/corr/abs-2108-00279,rude2004language}. Happy moments related to sleep are also more prevalent than those of the control category - poor sleep quality being one of the depression symptoms \cite{o2017might}. Subjects diagnosed with depression also have more happy moments related to music (e.g., \textit{"Together we crafted a legacy of music that has given so many people happiness and great memories."}, \textit{"I read a bit from my book and listened to music."}), which is indicative of the positive effects music therapy has on mental well-being \cite{tang2020effects}.

A surprising result was to find the death category from LIWC in the reasons for the happy moments of the control group. However, a closer manual analysis of the moments containing keywords from the death category revealed that, even if this category is comprised of death-related words (e.g., kill, dead, dying), they may not be used with their primary sense (e.g., \textit{"I have been able to increase dead lifts from 70 kg to 100kg."}) or the words may be used in the context of video games (e.g., \textit{"I killed 5 King Black Dragons."}).

To further understand the differences in the happy moments reported by users with depression and control, we compute the keyness score \cite{kilgarriff2009simple,gabrielatos2018keyness} for the two corpora. We perform a keyness analysis by comparing the frequencies of words from the happy moments of users diagnosed with depression (target corpus) to the frequencies of words from the texts of control users (reference corpus). We perform this analysis on nouns and verbs. We present the top 20 nouns and verbs from the two corpora (depression and control) ordered by their log-likelihood ratio (G$^2$) \cite{dunning1993accurate} in Figure \ref{fig:keyness-nouns} and \ref{fig:keyness-verbs}.

In Figure \ref{fig:keyness-nouns} we show the most frequent nouns in the happy moments of individuals diagnosed with depression in comparison with the control group. A similar pattern of words related to family and friends is also found in the keyness analysis of words from the happy moments of users from the depression group.

In Figure \ref{fig:keyness-verbs} we present the most frequent 20 verbs from each group. Users diagnosed with depression use more verbs related to emotions (e.g., feel, cry, cuddle), as opposed to users from the control group, which use more action verbs (e.g., build, lead, run) in the texts of their happy moments.

\section{Conclusion}
In this work, we explored the reasons for happiness in social media text. Using the HappyDB dataset of happy moments and the Positive-Unlabeled Learning (PU-learning) framework, we automatically constructed a model to extract happy moments from the eRisk dataset. Our linguistic analysis through LIWC and keyness information showed that people diagnosed with depression more commonly experience joy from moments related to their family and friends and attribute greater significance to such events, while control subjects have happiness moments related to common everyday activities such as subjects related to money and entertainment. 

This preliminary work is an exploration into manifestations of happiness of depressed individuals. The datasets used may be positively biased towards depressed individuals, which themselves attribute great value to their experiences of happiness, and choose to express it online; it is not clear the extent to which control users are expressing their happy moments online, rather than just enjoying them in the real world. However, our preliminary results are worthy of further development in future work, being in line with research performed in psychological studies \cite{tang2020effects,rude2004language}, and could prove helpful in developing therapeutic chatbots and developing suitable positive psychological interventions. Furthermore, we believe that our results pave the way to a more in-depth analysis of expressions of happiness, by analyzing aspects of agency and sociality in happy moments. Moreover, due to computational limitations, we trained our PU-learning model on a sub-sample of the corpus; in future work, we aim to include the entire HappyDB dataset into the training data and perform our analyses on larger datasets (i.e. RSDD \cite{yates-etal-2017-depression}).

% \nocite{*}
\section{Bibliographical References}\label{reference}
%\label{main:ref}

\bibliographystyle{lrec2022-bib}
\bibliography{refs}

% \section{Language Resource References}
% \label{lr:ref}
% \bibliographystylelanguageresource{lrec2022-bib}
% \bibliographylanguageresource{languageresource}

\end{document}